\documentclass[letterpaper, 10 pt, conference]{ieeeconf} 
\IEEEoverridecommandlockouts
\usepackage{cite}
\usepackage{notoccite}
\usepackage{amsmath,amssymb,amsfonts}
\usepackage{algorithmic}
\usepackage{graphicx}
\usepackage{textcomp}
\usepackage{xcolor}

\usepackage[bookmarks=false]{hyperref}

\let\labelindent\relax
\usepackage{enumitem}
\setlist[itemize]{noitemsep, nolistsep}
\newcommand{\subparagraph}{}
\usepackage{titlesec}
\usepackage[font=footnotesize]{caption}
\titlespacing{\section}{0pt}{*0.7}{*0.7}
\titlespacing{\subsection}{5pt}{*0.65}{*0.65}
\title{\LARGE \bf
Strawberry Detection using Mixed Training on Simulated and Real Data
}

\author{Sunny Goondram, Akansel Cosgun and Dana Kuli\'c \\Monash University, Australia
}

\begin{document}
\maketitle
\begin{abstract}
This paper demonstrates how simulated images can be useful for object detection tasks in the agricultural sector, where labeled data can be scarce and costly to collect. We consider training on mixed datasets with real and simulated data for strawberry detection in real images. Our results show that using the real dataset augmented by the simulated dataset resulted in slightly higher accuracy.

\end{abstract}

\section{Introduction}
\vspace{-0.1cm}

\begin{figure}[t!]
\centering
\includegraphics[trim={7cm 0cm 0cm 0cm}, clip, width=0.49\columnwidth]{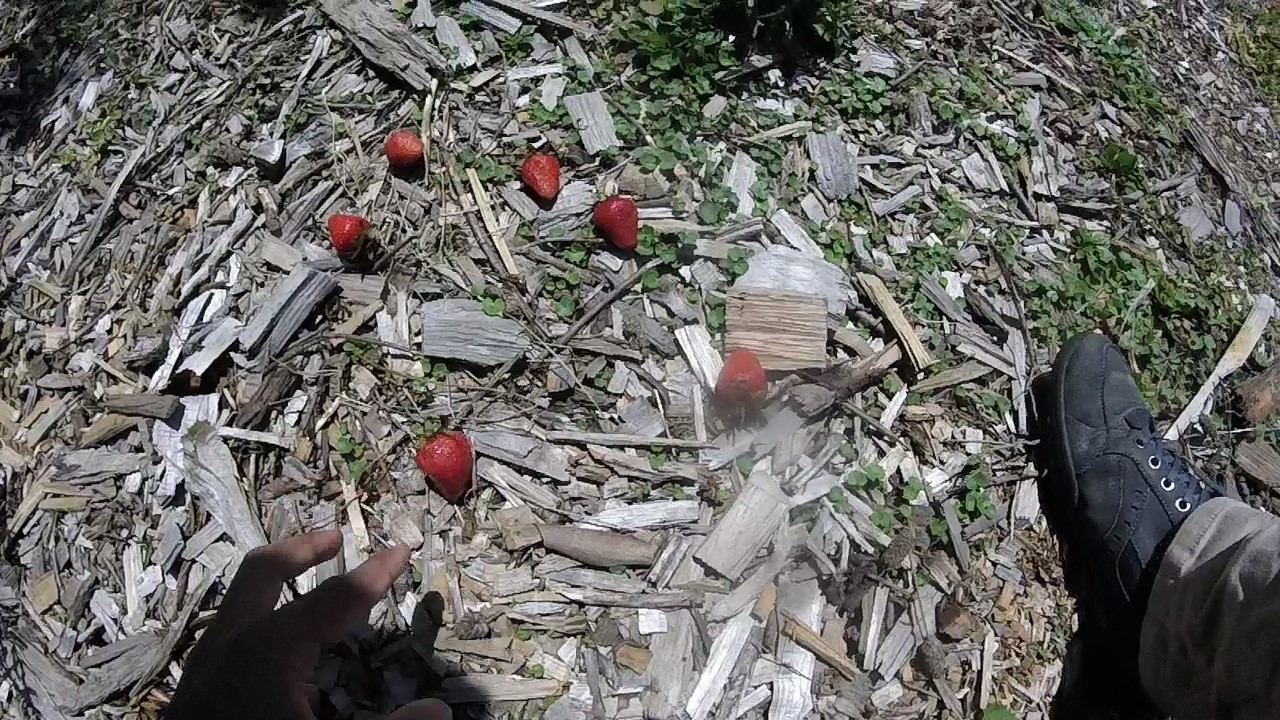}
\includegraphics[trim={7cm 0cm 0cm 0cm}, clip, width=0.49\columnwidth]{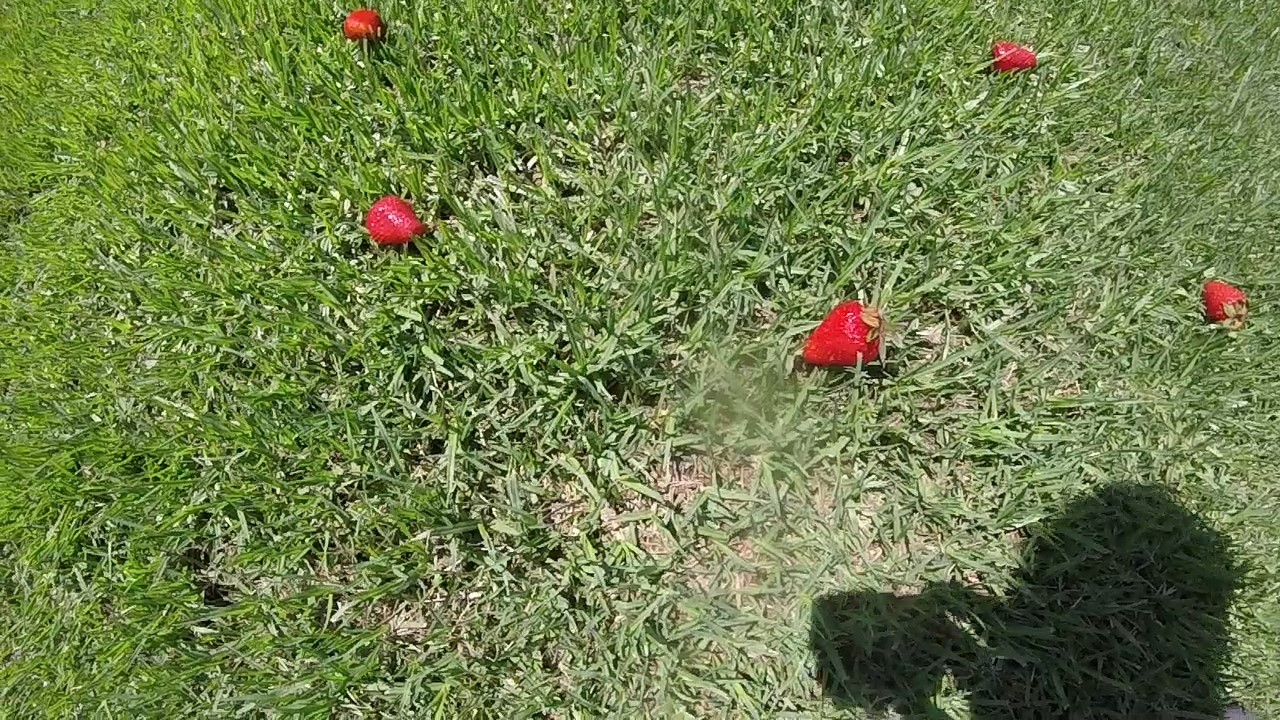}
\includegraphics[trim={0cm 0cm 0cm 5cm}, clip, width=0.49\columnwidth]{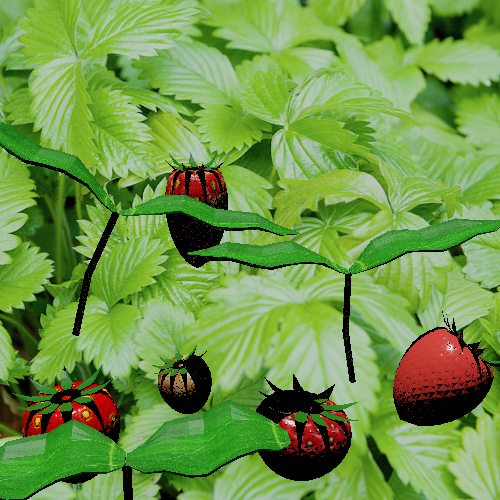}
\includegraphics[trim={0cm 0cm 0cm 5cm}, clip, width=0.49\columnwidth]{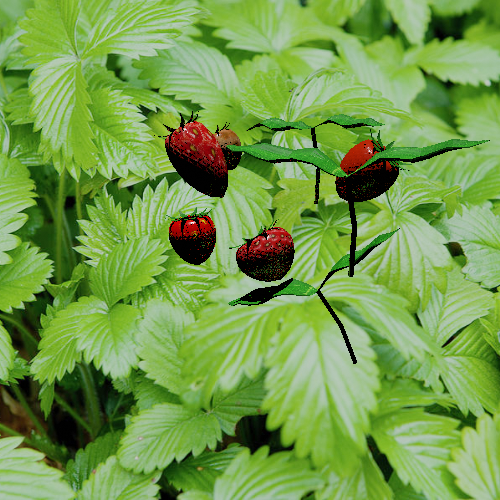}
\caption{Examples images from our strawberry datasets from the real world (top row) and simulated datasets (bottom row). }
\label{fig:fig1}
\end{figure}

Agriculture is a sector where a significant portion of the costs stem from manual labor, which encourages considerable research in developing automation for various aspects of agriculture and farming. Technological advances allow some crops such as potatoes to be mechanically harvested, however many crops, such as strawberries, still require human labour to be harvested. One possible reason for this lack of automation is the inherent difficulty in autonomously detecting and harvesting the delicate crops. A computer vision system that will automatically detect such fruits is a good first step towards automating the harvesting process. The advent of deep learning \cite{goodfellow2016deep} has enabled many applications in computer vision \cite{krizhevsky2012imagenet}, particularly in object detection \cite{ren2015faster, redmon2018yolov3}. The use of deep learning can be very useful towards automation in the agricultural sector \cite{kamilaris2018deep}. The object detection framework has been successfully applied to detecting fruits in orchards \cite{sa2016deepfruits,bargoti2017deep}. However, one major challenge is the lack of adequate datasets in agriculture. The collection and annotation of datasets can be costly and time-consuming, especially in the agriculture domain. Thanks to advancement of simulation tools, an alternative to real world data collection is the generation and use of simulated datasets. In contrast to real world data collection, simulated datasets offer advantages such as automatic labeling and the simulation of edge cases. The idea of using simulated data for training a neural network and then testing on real data has previously been explored in a fruit counting task \cite{rahnemoonfar2017deep}. Their model, however, is trained only on simulated data whereas in this paper we analyze training on mixed datasets with real and simulated data for strawberry detection in real images. We collect two RGB image datasets for the strawberry detection task: a real-world and a simulated dataset. Different combinations of the real and simulated datasets are then used to train an object detection neural network and the performance of these networks are evaluated using the real dataset. Some examples from the real and simulated dataset are shown in Fig. \ref{fig:fig1}.

\section{Data Collection}

\subsection{Real World}
To collect the real data for our project, a “concept strawberry farm” was created by buying ripe strawberries and placing them at different locations with different backgrounds. The strawberries were placed on the soil, near shrubs and on grass. 5 videos of these scenes were taken from a first-person view with a GoPro camera attached to the head of one person who moved around the strawberries and picked some of them up. The frames of these videos were extracted and the ground truth strawberry bounding boxes were manually annotated. A total of 824 images were collected with a resolution of 1280 by 720.

\subsection{Simulated}
We used Blender as a simulation modelling tool for generating the simulated dataset. A central base scene is created through modelling of five strawberries of different sizes and colors. Four of the strawberry models are ripe strawberries indicated by their red texture, whereas the unripe strawberry was green. The ripe strawberries has slightly different textures, to add more variation and realism. Leaves are also modelled so that in some scenes, the leaves can partly occlude the fruits. This is an important feature as, during strawberry picking, the strawberries are often occluded by the foliage. To create various scenes, we use an approach similar to \cite{ward2018deep}. We vary 3 aspects for creating each image: 

\begin{enumerate}
\item Fruit Pose: For each of the 5 strawberry models, a random position and orientation was chosen within a range.
\item Scene lighting: For each scene, it can be captured using 3 different lighting mode: a strong central one, a moderate central one and a moderate side one. 
\item Camera Pose: Once the scene is generated, we capture the scene from several viewpoints. After creating a scene through the fruit pose variation within a range, the strawberries are constrained to be within a region at the center of the scene. 5 viewpoints (camera poses) are manually determined such that the camera looks at the center of the region from different yaw angles around the vertical axis.
\end{enumerate}

The ground truth labels were generated automatically for the object detection task, in the forms of bounding boxes. Some examples of generated images are shown in Fig. \ref{fig:fig1} (bottom row). We collected a total of 3500 simulated images for our experiments, each with a resolution of 500 by 500, even though more data could be generated with different configurations.


\section{Experiments}

\subsection{Object Detection}
We use YOLOv3 \cite{redmon2018yolov3} for object detection, which is a popular object detection algorithm based on convolutional layers and residual connections. We chose YOLOv3 as it has a relatively higher inference speed for a comparable detection accuracy to the state-of-the-art.

\subsection{Mixed Training}

4 different training and testing sets are formed from real and simulated data:

\begin{itemize}
\item \textbf{Real only: } 700 real training and 124 real testing images
\item \textbf{Half real half sim:} 700 training (350 real and 350 simulated) and 124 real testing images. 
\item \textbf{5x more sim: } 4200 training (700 real and 3500 simulated) and 124 real testing images.
\item \textbf{Sim only: } 700 simulated training images and 124 real testing images.
\end{itemize}

All images are resized to 416 by 416 before being passed onto the network. Each training is done two times using a shuffle split cross-validation approach, where the training and testing data is chosen randomly from the pool of images of the selected dataset.

\subsection{Metrics}

For a detection threshold, precision refers to the degree with which the predictions are correct and recall refers to the degree with which the algorithm is able to find all the objects of interest. Average precision refers to the average of precision values for different detection thresholds. F1-score refers to the harmonic mean of the precision and recall. We compute precision, recall and F1-score as follows: 
\begin{equation}
\centering
precision = \frac{TP}{TP+FP},  recall = \frac{TP}{TP+FN}
\label{eq1}
\end{equation}
\begin{equation}
F1\_score = \frac{2*precision*recall}{precision+recall}
\label{eq3}
\end{equation}
where TP,FP and FN refer to number of true positive, false positive and false negative samples respectively.

\subsection{Results}
The strawberry detection results are shown in Table \ref{table:1}.
\begin{table}[ht!]
\begin{center}
\begin{tabular}{ |c|c|c|} 
\hline
Training Set & {F1 score(\%)} & {Average Precision(\%)}\\
\hline
{Real only} & 97.8 & \textbf{99.3} \\ 
\hline
{Half real half sim} & 97.6 & 98.2 \\ 
\hline
{5x more sim} & \textbf{98.3} & \textbf{99.3} \\ 
\hline
{Sim only} & 7.7 & 3 \\
\hline
\end{tabular}
\end{center}
\caption{Strawberry object detection results for each training set. Average F1 score(\%) and precision Precision(\%) is reported}
\label{table:1}
\end{table}

Training with \textbf{5x more sim} achieved the best F1 score with 98.3\%, which was slightly better than \textbf{Real only} (97.8\%) and \textbf{Half real half sim} (97.6\%). \textbf{Real only} and \textbf{5x more sim} both achieved the best average precision with 99.3\%. The results show that boosting the real dataset with a simulated dataset resulted in slightly higher accuracy as the F1 score is higher (same average precision). This implies that the features presented in the simulated data can reinforce the features presented in the real data.  The slight improvement of \textbf{5x more sim} over \textbf{Real only} justifies the use of additional simulated data whenever available.

Replacing half of the real dataset with simulated data resulted in comparable performance to the real only dataset, which shows that the simulated data was an adequate substitute to real images for this application. Training on only simulated images performed worse than any other approach, possibly because of the reality gap. 

\section{Conclusion}

We present a tool for generating simulated images for the strawberry detection task, as well as an analysis of using mixed training on simulated and real data. Our results show that using the real dataset augmented by the simulated dataset resulted in slightly higher object detection accuracy, and that simulated data could be used as a substitute to real. However, our analysis serves as an exploratory study and should be conducted with larger datasets and different domains. Future work includes creating more realistic simulations, collecting more real images from an actual strawberry farm, testing the approach on different fruits and exploring alternative approaches in sim-to-real transfer learning \cite{isele2017transferring}.

\bibliographystyle{IEEEtran}
\bibliography{references}

\begin{thebibliography}{10}
\providecommand{\url}[1]{#1}
\csname url@samestyle\endcsname
\providecommand{\newblock}{\relax}
\providecommand{\bibinfo}[2]{#2}
\providecommand{\BIBentrySTDinterwordspacing}{\spaceskip=0pt\relax}
\providecommand{\BIBentryALTinterwordstretchfactor}{4}
\providecommand{\BIBentryALTinterwordspacing}{\spaceskip=\fontdimen2\font plus
\BIBentryALTinterwordstretchfactor\fontdimen3\font minus
  \fontdimen4\font\relax}
\providecommand{\BIBforeignlanguage}[2]{{%
\expandafter\ifx\csname l@#1\endcsname\relax
\typeout{** WARNING: IEEEtran.bst: No hyphenation pattern has been}%
\typeout{** loaded for the language `#1'. Using the pattern for}%
\typeout{** the default language instead.}%
\else
\language=\csname l@#1\endcsname
\fi
#2}}
\providecommand{\BIBdecl}{\relax}
\BIBdecl

\bibitem{goodfellow2016deep}
I.~Goodfellow, Y.~Bengio, A.~Courville, and Y.~Bengio, \emph{Deep
  learning}.\hskip 1em plus 0.5em minus 0.4em\relax MIT press Cambridge, 2016,
  vol.~1.

\bibitem{krizhevsky2012imagenet}
A.~Krizhevsky, I.~Sutskever, and G.~E. Hinton, ``Imagenet classification with
  deep convolutional neural networks,'' in \emph{Advances in neural information
  processing systems}, 2012, pp. 1097--1105.

\bibitem{ren2015faster}
S.~Ren, K.~He, R.~Girshick, and J.~Sun, ``Faster r-cnn: Towards real-time
  object detection with region proposal networks,'' in \emph{Advances in neural
  information processing systems}, 2015, pp. 91--99.

\bibitem{redmon2018yolov3}
J.~Redmon and A.~Farhadi, ``Yolov3: An incremental improvement,'' \emph{arXiv
  preprint arXiv:1804.02767}, 2018.

\bibitem{kamilaris2018deep}
A.~Kamilaris and F.~X. Prenafeta-Bold{\'u}, ``Deep learning in agriculture: A
  survey,'' \emph{Computers and electronics in agriculture}, vol. 147, pp.
  70--90, 2018.

\bibitem{sa2016deepfruits}
I.~Sa, Z.~Ge, F.~Dayoub, B.~Upcroft, T.~Perez, and C.~McCool, ``Deepfruits: A
  fruit detection system using deep neural networks,'' \emph{Sensors}, vol.~16,
  no.~8, p. 1222, 2016.

\bibitem{bargoti2017deep}
S.~Bargoti and J.~Underwood, ``Deep fruit detection in orchards,'' in
  \emph{2017 IEEE International Conference on Robotics and Automation
  (ICRA)}.\hskip 1em plus 0.5em minus 0.4em\relax IEEE, 2017, pp. 3626--3633.

\bibitem{rahnemoonfar2017deep}
M.~Rahnemoonfar and C.~Sheppard, ``Deep count: fruit counting based on deep
  simulated learning,'' \emph{Sensors}, vol.~17, no.~4, p. 905, 2017.

\bibitem{ward2018deep}
D.~Ward, P.~Moghadam, and N.~Hudson, ``Deep leaf segmentation using synthetic
  data,'' \emph{arXiv preprint arXiv:1807.10931}, 2018.

\bibitem{isele2017transferring}
D.~Isele and A.~Cosgun, ``Transferring autonomous driving knowledge on
  simulated and real intersections,'' \emph{arXiv preprint arXiv:1712.01106},
  2017.

\end{thebibliography}
\end{document}